\documentclass[aps,prl,twocolumn,groupedaddress,superscriptaddress,showpacs]{revtex4-1}

\usepackage{hyperref}       % hyperlinks
\usepackage{url}            % simple URL typesetting
\usepackage{booktabs}       % professional-quality tables
\usepackage{amsfonts}       % blackboard math symbols
\usepackage{amssymb}
\usepackage{mathtools}
\usepackage{nicefrac}       % compact symbols for 1/2, etc.
\usepackage{microtype}      % microtypography
\usepackage{amsmath}
\usepackage{xspace}
\usepackage{xcolor,colortbl}
\usepackage{hhline}
\usepackage{bbm}
\usepackage{changes}
\usepackage{tabularx}
\usepackage{amsthm}
\usepackage[nameinlink]{cleveref}
\usepackage{changes}
\usepackage{tikz}
\usepackage{color,soul}
\usepackage{multirow}

\newcommand\tv[1]{{\pmb{\MakeLowercase{#1}}}}
\newcommand{\norm}[1]{\left\lVert#1\right\rVert}
\newcommand{\target}[1]{{{{#1}^{*}}}}

\def\eg{\emph{e.g.}}
\def\ie{\emph{i.e.}}

\begin{document}
\title{Implicit energy regularization of neural ordinary-differential-equation control}
\date{\today}
\author{Lucas B\"ottcher}
\email{lucasb@ucla.edu}
\affiliation{Computational Medicine, University of California, Los Angeles, 90095-1766, Los Angeles, United States}
\author{Nino Antulov-Fantulin}
\email{anino@ethz.ch}
\affiliation{Computational Social Science, ETH Zurich, 8092, Zurich, Switzerland}
\author{Thomas Asikis}
\email{asikist@ethz.ch}
\affiliation{Computational Social Science, ETH Zurich, 8092, Zurich, Switzerland}
\date{\today}
\begin{abstract}
Although optimal control problems of dynamical systems can be formulated within the framework of variational calculus, their solution for complex systems is often analytically and computationally intractable. In this Letter we present a versatile neural ordinary-differential-equation control (NODEC) framework with implicit energy regularization and use it to obtain neural-network-generated control signals that can steer dynamical systems towards a desired target state within a predefined amount of time. We demonstrate the ability of NODEC to learn control signals that closely resemble those found by corresponding optimal control frameworks in terms of control energy and deviation from the desired target state. Our results suggest that NODEC is capable to solve a wide range of control and optimization problems, including those that are analytically intractable.
\end{abstract}
\maketitle
The problem of how to optimally control complex systems has its roots in dynamical systems and optimization theory~\cite{kalman1960contributions,hautus1969controllability,lin1974structural,liu2016control}. Mathematically, a dynamical system is said to be ``controllable'' if it can be steered from any initial state ${\tv x}_0$ to any target state $\target{\tv{x}}$ in finite time $T$. Controlling complex dynamical systems is relevant in many applications such as (i) development of efficient and robust near-term quantum devices~\cite{mabuchi2009continuous,dong2010quantum}, (ii) regulatory network control~\cite{gottgens2015regulatory} in cellular biology, (iii) power-grid management~\cite{schafer2018dynamically}, (iv) design of stable financial systems~\cite{delpini2013evolution}, and (v) epidemic management~\cite{choi2021optimal}. 
 
Historically, an early work by Kalman in the 1960s led to the formulation of an analytical condition for the controllability of linear systems based on the rank of the controllability matrix~\cite{kalman1960contributions}. An equivalent condition, the so-called Popov--Belevitch--Hautus test~\cite{hautus1969controllability}, characterizes controllability of a linear system via its eigenmodes. More recently, concepts from the framework of structural controllability~\cite{lin1974structural} have been used to control complex networks~\cite{liu2011controllability} with a minimum set of control inputs (\ie, driver nodes) that can be determined by identifying all unmatched nodes in a maximum matching problem. The direct application of this framework to general network controllability problems is, however, complicated by several factors~\cite{pasqualetti2014controllability}. First, finding the minimum set of driver nodes for an arbitrary network is NP hard~\cite{olshevsky2014minimal}. Second, the design of an appropriate control signal is not specified in~\cite{liu2011controllability} and its implementation may not be realizable in practice~\cite{pasqualetti2014controllability}. Third, in the presence of nodal self-dynamics, which were not included in the controllability framework~\cite{liu2011controllability}, a single time-varying input is sufficient to achieve structural controllability~\cite{cowan2012nodal}, challenging the findings of~\cite{liu2011controllability}.

The solution of general optimal control problems is based on two main approaches: (i) Pontryagin's maximum principle~\cite{mcshane1989calculus} (necessary condition),
which is a boundary-value problem in a Hamiltonian framework or 
(ii) solving the Hamilton–Jacobi–Bellman (HJB) partial-differential equation (necessary and sufficient condition)~\cite{zhou1990maximum}. Since the HJB equation usually does not admit smooth solutions~\cite{Frankowska}, different approximate dynamic programming methods are used~\cite{abu2005nearly, bellman2015applied}.  

In this Letter, we study the ability of neural ordinary-differential-equation control (NODEC)~\cite{asikis2020nnc} to steer different dynamical processes towards desired target states without explicitly accounting for an energy-regularization term in the corresponding loss function. 
The proposed NODEC framework extends the neural ODE formalism~\cite{chen2018neural} to general control problems and automatically learns control signals that steer the evolution of an underlying networked dynamical system. Using analytical and numerical arguments, we show why NODEC is able to closely resemble the control energy of optimal control through an implicit energy regularization, resulting from the interplay of neural network initizalization and an induced gradient descent.
\paragraph*{Neural-network control.}
Before outlining the basic principles of NODEC, we first provide a mathematical formulation of the control problem of networked dynamical systems. We consider a network that consists of $N$ nodes whose states are represented by the state vector $\tv{x}(t)\in \mathbb{R}^N$. Initially, nodes are in state $\tv{x}(0)$ and steered towards a target state $\target{\tv{x}}$ at time $T$ (\ie, $\tv{\tv{x}}(T)=\target{\tv{x}}$) by means of suitable control inputs. Interactions between nodes are described by the dynamical system
\begin{equation}
\dot{\tv{x}}(t) = f(\tv{x}(t),\tv{u}(t))
\label{eq:dynamical_system}
\end{equation}
and subject to the constraint that the control function $\tv{u}(t)\in \mathbb{R}^{M}$ minimizes the cost function
\begin{equation}
J = \int_0^T L(\tv{x}(t'),\tv{u}(t')) \, dt' + C(\tv{x}(T))\,. 
\label{eq:cost_function}
\end{equation}
The function $f\colon \mathbb{R}^N \rightarrow \mathbb{R}^N$ in Eq.~\eqref{eq:dynamical_system} accounts for both the interactions between nodes $1,\dots,N$ and the influence of external control inputs $\tv{u}(t)$ on the dynamics. Note that the number of control inputs $M$ is smaller than or equal to $N$. For linear systems, we describe node-node interactions and external control inputs by $f(\tv{x},\tv{u})=A \tv{x}+B \tv{u}$. The first term in Eq.~\eqref{eq:cost_function} is the integrated cost over the control horizon $T$, \eg, the control energy 
\begin{equation}
E_T[\tv{u}] = \int_{0}^{T} \norm{\tv{u}(t')}_2^2\, dt'
\label{eq:energy}
\end{equation}
if $L=\norm{\tv{u}(t')}_2^2$. $C(\tv{x}(T))$ is the final cost (or bequest value). Most common formulations of optimal control include the control-energy term \eqref{eq:energy} directly in the cost function~\cite{yan2012controlling,sun2013controllability}. This approach corresponds to an explicit minimization of the control energy.

In NODEC, we take a complementary approach to reach a desired target state $\target{\tv{x}}$ in finite time $T$ and proceed in two steps. First, we approximate and solve the dynamical system in terms of neural ODEs~\cite{chen2018neural}. In particular, we describe the control input $\tv{u}(t)$ by a neural network with weight vector $\tv{w}$ such that the corresponding control-input representation is $\hat{\tv{u}}(t;\tv{w})$. Second, we use a suitable loss function $J(\tv{x},\target{\tv{x}})$ and a gradient-descent algorithm to iteratively determine the weight vector $\tv{w}$ according to~\cite{asikis2020nnc}
\begin{equation}
\tv{w}^{(n+1)} = \tv{w}^{(n)} - \eta \nabla_{\tv{w}^{(n)}} J(\tv{x},\target{\tv{x}}),
\label{eq:gradient_descent}
\end{equation}
where the superscript indicates the current number of gradient-descent steps, and $\eta$ is the learning rate. For the loss function $J(\cdot)$, we use the mean-squared error
\begin{equation} 
J(\tv{x}(T),\target{\tv{x}})=\frac{1}{N}\norm{\tv{x}(T)-\target{\tv{x}}}^2_2.
\label{eq:loss_function}
\end{equation}
In order to calculate $\nabla_{\tv{w}^{(n)}}J(\cdot)$, we use automatic differentiation methods~\cite{baydin2018automatic}, where the gradients ``flow'' through the underlying neural network, that is time-unfolded~\cite{schafer2006recurrent} by ODE solvers~\cite{shampine2018numerical}. We will show in the following paragraphs that, even without including the energy cost \eqref{eq:energy} in the loss function \eqref{eq:loss_function}, NODEC approximates optimal control by minimizing the control energy \eqref{eq:energy}. All neural-network architectures, hyperparameters, and numerical solvers are reported in the Supplemental Information (SI). 
\paragraph*{Approximating optimal control.}
\begin{figure}
\includegraphics[width=0.49\textwidth]{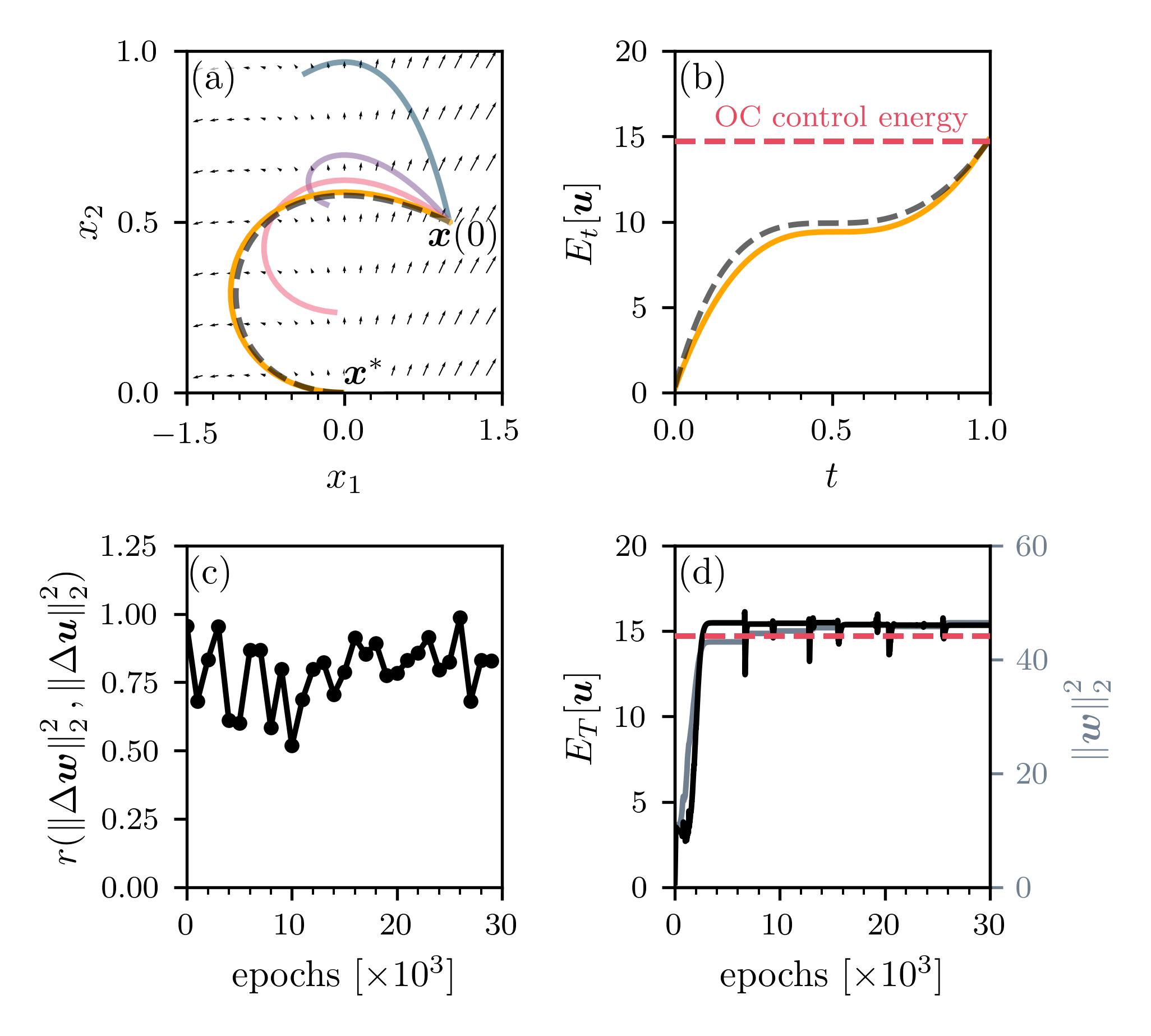}
 \caption{Neural ordinary-differential-equation control of a two-node system. (a) Different neural-network control trajectories for $\tv{x}(0)=(1,0.5)^{\rm T}$, $\target{\tv{x}}=(0,0)^{\rm T}$, and $T=1$ after 500 (blue), 1500 (purple), 2000 (red), and 30000 (orange) training epochs with learning rate $\eta=0.02$. The dashed black line is the corresponding optimal-control trajectory and black arrows indicate the vector field of the linear dynamical system $f(\tv{x},\tv{u}) = A \tv{x}+B\tv{u}$ with matrices $A$ and $B$ as in Eq.~\eqref{eq:linear_system} (b) Evolution of the control energy $E_t[\mathbf{u}]$ for NODEC after 30000 training epochs (solid orange line) and optimal control (dashed black line). (c) Correlations between squared norm differences of neural-network weights $\mathbf{w}$ and control inputs $\mathbf{u}$. (d) The control energy $E_T[\tv{u}]$ (black solid line) and squared norm of the neural-network weights $\mathbf{w}$ (solid grey line) of NODEC as a function of training epochs. In (c,d), we indicate the total OC control energy by a dashed red line.}
 \label{fig:nnc_optimal}
\end{figure}
We now compare the control performance of NODEC for linear systems (\ie, $f(\tv{x},\tv{u}) = A \tv{x}+B\tv{u}$), for which there exist analytical OC inputs~\cite{yan2012controlling}
\begin{equation}
\tv{u}^\ast(t) = B^{\top}e^{A(T - t)}{W(T)}^{-1}\tv{v}({T})
\label{eq:optimalcontrol}
\end{equation}
that minimize the control energy $E[\tv{u}(t)]$ [Eq.~\eqref{eq:energy}]. For the derivation of Eq.~\eqref{eq:optimalcontrol}, one applies Pontryagin's maximum principle to the Hamiltonian $H= \norm{\tv{u}(t)}_2^2 + \lambda^T(A \tv{x} + B \tv{u})$~\cite{yan2012controlling}, where $\lambda$ is an adjoint variable. The vector $\tv{v}({T})=\tv{x}({T})-e^{A T} \tv{x}_0$ in Eq.~\eqref{eq:optimalcontrol} is the difference between the target state $\tv{x}({T})$ and initial state $\tv{x}(0)$ under free evolution. The matrix $W(T)$ is the controllability Gramian and defined as
\begin{equation}
\label{eq:gramian}
W(T) = \int_0^{T} e^{A t} B B^\top e^{A^{\top} t}\, dt. 
\end{equation}
As an example of linear dynamics, we consider a two-state system with~\cite{yan2012controlling,sun2013controllability}
\begin{equation}
A = \begin{pmatrix}
1 & 0\\
1 & 0
\end{pmatrix}\quad \text{and}\quad 
B = \begin{pmatrix}
1\\
0
\end{pmatrix}\,.
\label{eq:linear_system}
\end{equation}
The control task is to steer the system from $\tv{x}(0)=(1,0.5)^{\rm T}$ to $\target{\tv{x}}=(0,0)^{\rm T}$ in finite time $T=1$.

In Fig.~\ref{fig:nnc_optimal}(a), we show NODEC trajectories after 500 (blue), 1000 (purple), 1500 (red), and 30000 (orange) training epochs together with an OC control trajectory (dashed black line). Note that the geodesic that connects $\tv{x}(0)$ and $\target{\tv{x}}(T)$ is not minimizing the control energy, because it would require large control inputs to steer the dynamics against the vector field [black arrows in Fig.~\ref{fig:nnc_optimal}(a)]. In alignment with the almost identical control trajectories of NODEC and OC, we also find that the energy evolution of NODEC almost perfectly coincides with that of OC [Fig.~\ref{fig:nnc_optimal}(b)], hinting at an implicit energy regularization of NODEC.
\paragraph*{Implicit energy regularization.}
To provide insights into the observed implicit energy regularization of NODEC [Fig.~\ref{fig:nnc_optimal}(b)], we show that a gradient descent in the neural-network weights $\tv{w}$ induces a gradient descent in the control input $\hat{\tv{u}}(t;\tv{w})$. 

The evolution of the state vector $\tv{x}(t)$ is described by Eq.~\eqref{eq:dynamical_system} and is a function of $\hat{\tv{u}}(t;\tv{w})$. We now expand $\hat{\tv{u}}(t;\tv{w}^{(n+1)})=\hat{\tv{u}}(t;\tv{w}^{(n)}+\Delta \tv{w}^{(n)})$ with $\Delta \tv{w}^{(n)}=-\eta\nabla_{\tv{w}^{(n)}}J$ for small $\Delta \tv{w}^{(n)}$ while keeping $t$ constant. This expansion yields
\begin{equation}
\hat{\tv{u}}(t;\tv{w}^{(n+1)}) = \hat{\tv{u}}(t;\tv{w}^{(n)}) + \mathcal{J}_{\hat{\tv{u}}}  \Delta \tv{w}^{(n)},
\end{equation}
where $\mathcal{J}_{\hat{\tv{u}}}$ is the Jacobian of $\hat{\tv{u}}$ with elements $(\mathcal{J}_{\hat{\tv{u}}})_{ij} = \partial \hat{\tv{u}}_i/\partial \tv{w}_j $. Note that we can make $\Delta \tv{w}^{(n)}$ arbitrarily small by using a small learning rate $\eta$.

Since $\Delta \tv{w}^{(n)} \propto \nabla_{\tv{w}^{(n)}}J $ and $\nabla_{\tv{w}^{(n)}}J =\mathcal{J}_{\hat{\tv{u}}}^T \nabla_{\hat{\tv{u}}} J
$, we obtain
\begin{equation}
\hat{\tv{u}}(t;\tv{w}^{(n+1)}) = \hat{\tv{u}}(t;\tv{w}^{(n)}) - \eta \mathcal{J}_{\hat{\tv{u}}} \mathcal{J}_{\hat{\tv{u}}}^T \nabla_{\hat{\tv{u}}} J.
\label{eq:grad_desc_u}
\end{equation}
According to Eq.~\eqref{eq:grad_desc_u}, a gradient descent in $\tv{w}$ [Eq.~\eqref{eq:gradient_descent}] may induce a gradient descent in $\hat{\tv{u}}$, where the square matrix $\mathcal{J}_{\hat{\tv{u}}} \mathcal{J}_{\hat{\tv{u}}}^T$ acts as a linear transformation on $\nabla_{\hat{\tv{u}}} J$. 

To better understand the implications of this result, we briefly summarize the control steps of NODEC. As described in the prior paragraphs and as illustrated in Fig.~\ref{fig:nnc_optimal}(a), NODEC starts with a small initial control signal $\hat{\tv{u}}^{(0)}(t;\tv{w}^{(0)})$, then integrates the dynamical system \eqref{eq:dynamical_system}, and performs a gradient descent in $\tv{w}$ according to Eq.~\eqref{eq:gradient_descent}. The closer the final state $\tv{x}(T)$ to the target state $\target{\tv{x}}$, the smaller the loss \eqref{eq:loss_function} and the change in $\tv{w}$ [and in $\hat{\tv{u}}$ due to Eq.~\eqref{eq:grad_desc_u}]. If we initialize NODEC with a sufficiently small control input and learning rate, it will produce control trajectories that follow the vector field of the dynamical system in a ``go-with-the-flow'' manner and slowly adapt $\hat{\tv{u}}$ to reach the desired target state. Because of the induced gradient descent \eqref{eq:grad_desc_u}, the resulting control approximates OC methods that minimize the control energy [see the comparison between the final control energy of OC and NODEC in Figs.~\ref{fig:nnc_optimal}(b,d)]. This way of controlling dynamical systems is markedly different from standard (optimal) control formulations~\cite{controlBook} that are, for instance, based on Pontryagin's maximum principle and require one to explicitly minimize the control energy by including $\norm{\tv{u}}_2^2$ in the Hamiltonian and solving the adjoint system~\cite{controlBook}. NODEC thus provides a complementary approach for solving general control problems.
\begin{figure}
    \centering
    \includegraphics[width = 0.49\textwidth]{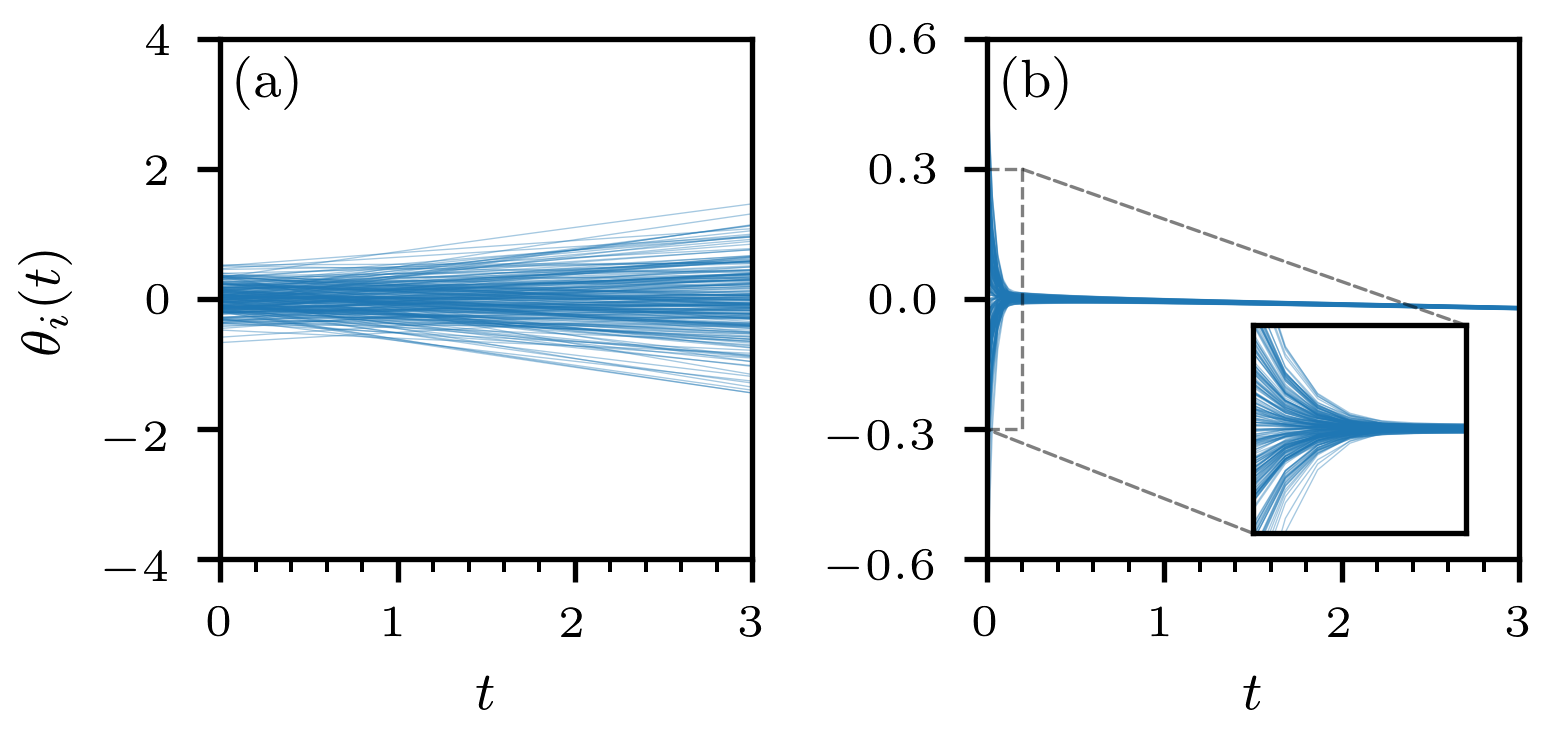}
    \caption{Synchronization of coupled oscillators. The evolution of oscillator phases $\theta_i(t)$ ($1\leq i \leq N$) in a complete network that consists of $N=225$ coupled Kuramoto oscillators [Eqs.~\eqref{eq:kuramoto} and \eqref{eq:kuramoto2}] with $K=0.1K^*$. All phases are initially distributed according to a normal distribution with mean $0$ and standard deviation $0.2$. (a) The control input is set to $u_i(t)=1$ for all $i$ (``uncontrolled dynamics''), leading to increasing phase differences over times. (b) NODEC synchronizes the system of coupled oscillators.}
    \label{fig:uncontrolled_vs_controlled}
\end{figure}
\begin{figure*}
    \centering
    \includegraphics[width = 0.84\textwidth]{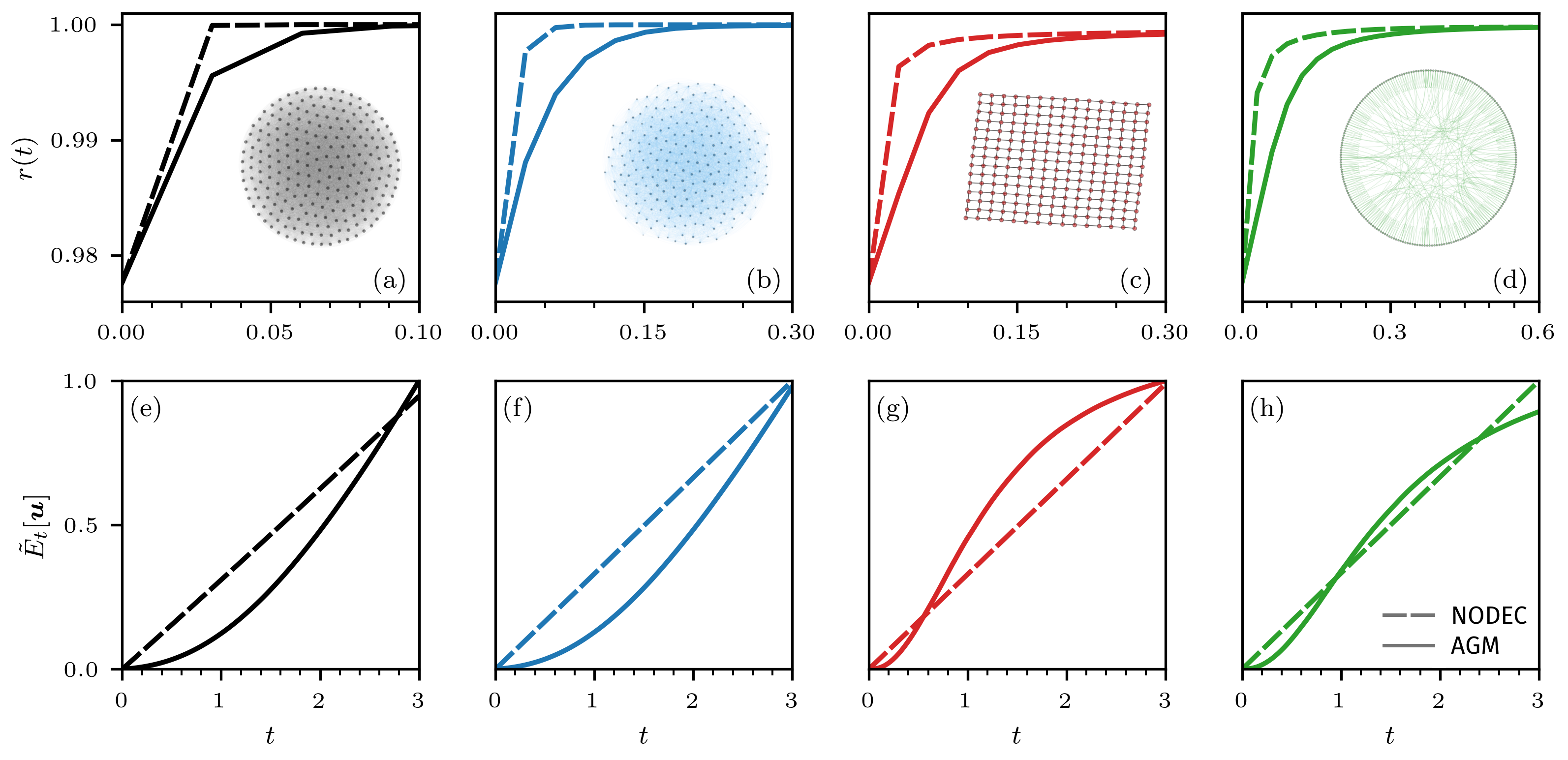}
    \caption{Controlling coupled oscillators with NODEC and the AGM. We test the performance of NODEC and AGM to control coupled Kuramoto oscillators on a (a,e) complete network (black lines), (b,f) Erd{\H o}s--R{\'e}nyi network $G(N,p)$ with $p=0.3$ (blue lines), (c,g) square lattice (red lines), and (d,h) Watts--Strogatz network with degree $k=5$ and rewiring probability $p=0.3$ (green lines). All graphs have $N=225$ nodes and the total simulation time is $T=3$. Panels (a--d) show the order parameter $r(t)$ and panels (e--h) show the control energy $\tilde{E}_t(\mathbf{u})=E_t(\mathbf{u})/\max(E_t^{\rm NODEC}(\mathbf{u}),E_t^{\rm AGM}(\mathbf{u}))$. Dashed and solid lines indicate NODEC and AGM solutions, respectively.}
    \label{fig:kuramoto_control}
\end{figure*}

The induced gradient descent \eqref{eq:grad_desc_u} can be directly observed in the positive correlations between $\norm{\Delta \tv{w}}_2^2=\norm{\tv{w}^{(n+1)}-\tv{w}^{(n)}}_2^2$ and $\norm{\Delta \tv{u}}_2^2=\norm{\tv{u}^{(n+1)}-\tv{u}^{(n)}}_2^2$ [Fig.~\ref{fig:nnc_optimal}(c)]. Black disks indicate correlation coefficients ($p < 10^{-9}$) that are each calculated for $10^3$ consecutive epochs and solid black lines are guides to the eye. After initializing NODEC for the linear two-state system \eqref{eq:linear_system} with weights that correspond to a small control input, we observe positive correlations between $ \norm{\Delta \tv{w}}_2^2$ and $ \norm{\Delta \tv{u}}_2^2$ with a large correlation coefficient of 0.96 for the first 1000 training epochs. The mean correlation coefficient is about $0.76$. Changes in the correlation behavior reflect different training stages that are necessary to capture the strong curvature in the OC control trajectory [dashed black line in Fig.~\ref{fig:nnc_optimal}(a)]. Between 1500 and 2000 training epochs, NODEC approximates the basic shape of the OC trajectory [solid red line and red disks in Fig.~\ref{fig:nnc_optimal}(a)] and then fine-tunes the weights $\tv{w}$ to match OC as closely as possible [solid orange line and orange disks in Fig.~\ref{fig:nnc_optimal}(a)]. The initial OC approximation phase that lasts up to about 2000 training epochs (before weight ``fine-tuning'') is also visible in the evolution of $\norm{\tv{w}}_2^2$ and $ \norm{\tv{u}}_2^2$ [Fig.~\ref{fig:nnc_optimal}(d)].

We again emphasize that the performance of NODEC and its induced-gradient descent mechanism depends on the choice of initial weights $\tv{w}^0$ [and thus $\tv{u}(t;\tv{w}^0)$]. The initialization that we use to obtain the results of Fig.~\ref{fig:nnc_optimal} is based on energy values which are distributed in the interval $\left[5, 7\right]$. These values are small enough for NODEC to let it move with the vector field of the underlying dynamical system and approximate OC.

After having outlined the mechanisms underlying the observed energy regularization of NODEC, we now turn towards non-linear systems.
\paragraph*{Neural ordinary-differential-equation control of Kuramoto oscillators.}
As an example of a non-linear system, we consider the Kuramoto model~\cite{kuramoto1975self}, which describes coupled oscillators with phases $\theta_i$ and intrinsic frequencies $\omega_i$ ($1\leq i \leq N$) according to~\cite{kuramoto1975self}
\begin{align}
\begin{split}
\dot{\Theta}(t) &= \Omega + f(\Theta(t),u(t)),\\
\Theta(0)&=\Theta_0,
\label{eq:kuramoto}
\end{split}
\end{align}
where $\Theta=(\theta_1,\dots,\theta_N)^\top$ and $\Omega=(\omega_1,\dots,\omega_N)^\top$. In our following numerical experiments, we use natural frequencies and initial phases that are normally-distributed with mean $0$ and standard deviation $0.2$. Interactions between oscillators and the influence of control inputs $u_i(t)$ on oscillator $i$ are modeled via
\begin{equation}
f_i(\Theta(t),u(t))=\frac{K u_i(t)}{N} \sum_{j=1}^N A_{ij} \sin(\theta_j(t)-\theta_i(t)),
\label{eq:kuramoto2}
\end{equation}
where $K$ is the coupling strength and $A_{ij}$ are the adjacency matrix components of the underlying (undirected) network. As a measure of synchronization at the final time $T$, we use the complete synchronization condition~\cite{ha2016emergence,biccari2020stochastic}
\begin{equation}
|\dot{\theta}_i(T)-\dot{\theta}_j(T)|=0~\text{for}~(i,j)\in E,
\label{eq:complete_synchronization}
\end{equation}
where $E$ is the set of edges. If Eq.~\eqref{eq:complete_synchronization} is satisfied, all connected oscillators have constant phase differences. For unit control inputs (\ie, $u_i(t)=1$), the system \eqref{eq:kuramoto} has a unique and stable synchronized state if the coupling constant exceeds a critical value~\cite{dorfler2013synchronization}
\begin{equation}
K^*=\norm{L^\dagger \Omega}_{E,\infty}, 
\end{equation}
where $L^\dagger$ is the pseudo-inverse of the corresponding graph Laplacian and $\norm{\tv{x}}_{E,\infty}=\max_{(i,j)\in E}|x_i-x_j|$ is the maximum distance between elements in $\tv{x}=(x_1,\dots,x_N)^\top$ that are connected via an edge in $E$. In all numerical simulations, we set $K=0.1 K^*$ such that control inputs $u_i(t)>1$ are needed to synchronize the system.

For a global control $u(t)$ (\ie, $u_i(t)=u(t)$ for all $i$), there exists an OC input $u^*(t)$ satisfying~\cite{biccari2020stochastic}
\begin{align}
 \target{u} &= \min_{u} J(u) \\
J(\Theta,u)&=\frac{1}{2}\sum_{i,j} A_{ij} \sin^2(\theta_j(T)-\theta_i(T))+\frac{\beta}{2} E[u] \label{eq:loss_kuramoto}, 
\end{align}
where the parameter $\beta$ determines the influence of the energy regularization term. Note that minimizing $J(\Theta,u)$ is consistent with Eq.~\eqref{eq:complete_synchronization} since
\begin{equation}
\sin(\theta_j(T)-\theta_i(T)) = 0 \implies \theta_j(T)-\theta_i(T) = k \pi,~k \in \mathbb{Z}.
\end{equation}
An optimal control for the outlined non-linear control problem, the so-called the adjoint-gradient method (AGM), can be derived using Pontryagin's maximum principle and a gradient descent in $u$~\cite{biccari2020stochastic}:
\begin{equation}
u^{(n+1)}=u^{(n)}-\tilde{\eta} \left[\beta u^{(n)} + \frac{K}{N} \sum_{i=1}^N \lambda_i \sum_{j=1}^N A_{ij} \sin(\theta_j-\theta_i) \right],
\label{eq:optimal_kuramoto}
\end{equation}
where $\tilde{\eta}$ is the learning rate and $\tv{\lambda}=(\lambda_1,\dots,\lambda_N)^\top$ is the solution of the adjoint system
\begin{equation}
-\dot{\lambda}_i = -\frac{K u \lambda_i}{N} \sum_{i\neq j} A_{ij} \cos(\theta_j-\theta_i)+\frac{K u}{N} \sum_{i\neq j} A_{ij}\lambda_j\cos(\theta_j-\theta_i),
\label{eq:AGM}
\end{equation}
and $\lambda_i(T)=1/2\sum_{i \neq j} A_{ij} \sin(2\theta_i(T)-2\theta_j(T))$.

We compare the control performance of NODEC, which solves Eq.~\eqref{eq:kuramoto} using neural ODEs, with that of the AGM for a global control function. Note that NODEC learns $\hat{u}^*(t;\tv{w})$ based on the loss function \eqref{eq:loss_kuramoto} \emph{without} energy regularization term and a gradient descent in $\tv{w}$. All employed network architectures and training parameters are summarized in the SI and in \cite{repository}. 

For a complete graph with $N=225$ nodes and $T=3$, we show a system of uncontrolled oscillators with $u_i(t)=1$ for all $i$ in Fig.~\ref{fig:uncontrolled_vs_controlled}(a). As shown in Fig.~\ref{fig:uncontrolled_vs_controlled}(b), NODEC can learn control inputs that drive the system of coupled oscillators into a synchronized state. To quantify the degree of synchronization, we use the order parameter $r(t) =N^{-1}\sqrt{\sum_{i,j}\cos \left[\theta_j(t)-\theta_i(t)\right]}$~\footnote{Here we used that the square of the magnitude of the complex order parameter $z=r e^{i \psi(t)}=N^{-1} \sum_{j=1}^N e^{i \theta_j(t)}$~\cite{kuramoto1975self} can be expressed as $r(t)^2=|z|^2=N^{-2} \sum_{i,j} e^{i (\theta_j(t)-\theta_i(t))}=N^{-2} \sum_{i,j}\cos\left[\theta_j(t)-\theta_i(t)\right]$.}. A value of $r(t)=1$ indicates that all oscillators have the same phase. In Fig.~\ref{fig:kuramoto_control} we show the evolution of the order parameter $r(t)$ and control energy $E_t[\mathbf{u}]$ for AGM (solid lines) and NODEC (dashed lines). We study the control performance of both methods on a complete graph (black lines), Erd{\H o}s--R{\'e}nyi network $G(N,p)$ with $p=0.3$ (blue lines), square lattice (red lines), and Watts--Strogatz network with degree $k=5$ and rewiring probability $p=0.3$ (green lines). All networks consist of $N=225$ oscillators. In the SI, we also demonstrate NODEC's ability to control Kuramoto dynamics in a square lattice with $10^4$ nodes.

For all networks, we observe that NODEC reaches synchronization slightly faster than the AGM [Fig.~\ref{fig:kuramoto_control}(a--d)]. We optimized the hyperparameters (\eg, the number of training epochs) of the neural network underlying NODEC such that the control energy and degree of synchronization lie in a similar range to those of the AGM [Fig.~\ref{fig:kuramoto_control}(e--h)]. Our results thus indicate that NODEC is able to achieve control energies similar to those of OC also for non-linear networked dynamics.
\paragraph*{Concluding remarks.}
We used NODEC, a control framework that is based on neural ODEs, to steer linear and non-linear networked dynamical systems into desired target states. For the considered linear dynamics, we compared NODEC with the corresponding analytical optimal-control solution and found that NODEC is not only able to drive the dynamical system into a desired target state, but also is able to approximate the optimal-control energy. We supported this observation with analytical arguments and further compared NODEC with an optimal-control method for synchronizing oscillators in different networks, again showing that NODEC is able to approximate the optimal-control energy. Neural ODE control frameworks are very versatile, complement existing control approaches, and may be useful to solve complex and analytically intractable control problems.
\acknowledgements
LB acknowledges financial support from the SNF (P2EZP2\_191888). LB and TA contributed equally to this work. All source codes and neural-network architectures are publicly available at~\cite{repository}.
\bibliography{refs.bib}
\end{document}